# Quantum Annealing for Clustering

**Kenichi Kurihara**
Google,
Tokyo, Japan

**Shu Tanaka**
Institute for Solid State Physics,
University of Tokyo
Chiba, Japan

**Seiji Miyashita**
Dept. of Physics,
University of Tokyo, Tokyo, Japan
CREST, Saitama, Japan

## Abstract

This paper studies quantum annealing (QA) for clustering, which can be seen as an extension of simulated annealing (SA). We derive a QA algorithm for clustering and propose an annealing schedule, which is crucial in practice. Experiments show the proposed QA algorithm finds better clustering assignments than SA. Furthermore, QA is as easy as SA to implement.

## 1 Introduction

Clustering is one of the most popular methods in data mining. Typically, clustering problems are formulated as optimization problems, which are solved by algorithms, for example the EM algorithm or convex relaxation. However, clustering is typically NP-hard. The simulated annealing (SA) (Kirkpatrick et al., 1983) is a promising candidate. Geman and Geman (1984) proved SA was able to find the global optimum with a slow cooling schedule of temperature $T$. Although their schedule is in practice too slow for clustering of a large amount of data, it is well known that SA still finds a reasonably good solution even with a faster schedule than what Geman and Geman proposed.

In statistical mechanics, quantum annealing (QA) has been proposed as a novel alternative to SA (Kadowaki and Nishimori, 1998; Santoro et al., 2002; Matsuda et al., 2009). QA adds another dimension, $\Gamma$, to SA for annealing, see Fig.1. Thus, it can be seen as an extension of SA. QA has succeeded in specific problems, e.g. the Ising model in statistical mechanics, and it is still unclear that QA works better than SA in general. We do not actually think QA intuitively helps clustering, but we apply QA to clustering just as procedure to derive an algorithm. A derived QA algorithm depends on the definition of quantum effect $\mathcal{H}_{\mathrm{q}}$. We propose quantum effect $\mathcal{H}_{\mathrm{q}}$, which leads to a search strategy fit to clustering. Our contribution is,

1. to propose a QA-based optimization algorithm for clustering, in particular
   (a) quantum effect $\mathcal{H}_{\mathrm{q}}$ for clustering
   (b) and a good annealing schedule, which is crucial for applications,
2. and to experimentally show the proposed algorithm optimizes clustering assignments better than SA.

We also show the proposed algorithm is as easy as SA to implement.

The algorithm we propose is a Markov chain Monte Carlo (MCMC) sampler, which we call QA-ST sampler. As we explain later, a naive QA sampler is intractable even with MCMC. Thus, we approximate QA by the Suzuki-Trotter (ST) expansion (Trotter, 1959; Suzuki, 1976) to derive a tractable sampler, which is the QA-ST sampler. QA-ST looks like parallel $m$ SAs with interaction $f$ (see Fig.2). At the beginning of the annealing process, QA-ST is almost the same as $m$ SAs. Hence, QA-ST finds $m$ (local) optima independently. As the annealing process continues, interaction $f$ in Fig.2 becomes stronger to move $m$ states closer. QA-ST at the end picks up the state with the lowest energy in $m$ states as the final solution.

QA-ST with the proposed quantum effect $\mathcal{H}_{\mathrm{q}}$ works well for clustering. Fig.3 is an example where data points are grouped into four clusters. $\sigma_1$ and $\sigma_2$ are locally optimal and $\sigma^*$ is globally optimal. Suppose $m$ is equal to two and $\sigma_1$ and $\sigma_2$ in Fig.2 correspond to $\sigma_1$ and $\sigma_2$ in Fig.3. Although $\sigma_1$ and $\sigma_2$ are local optima, the interaction $f$ in Fig.2 allows $\sigma_1$ and $\sigma_2$ to search for a better clustering assignment between $\sigma_1$ and $\sigma_2$. Quantum effect $\mathcal{H}_{\mathrm{q}}$ defines the distance metric of clustering assignments. In this case, the proposed $\mathcal{H}_{\mathrm{q}}$ locates $\sigma^*$ between $\sigma_1$ and $\sigma_2$. Thus, the interaction $f$ gives good chance to go to $\sigma^*$ because $f$ makes $\sigma_1$ and $\sigma_2$ closer (see Fig.2). The proposed algorithm actually finds $\sigma^*$ from $\sigma_1$ and $\sigma_2$. Fig.3 is just an example. However, a similar situation often occurs in clustering. Clustering algorithms in most cases give "almost" globally optimal solutions like $\sigma_1$ and $\sigma_2$, where the majority of data points are well-clustered, but some of them are not. Thus, a better clustering assignment can be constructed by picking up well-clustered data points from many sub-optimal clustering assignments. Note an assignment constructed in such a way is located between the sub-optimal ones by the proposed quantum effect $\mathcal{H}_{\mathrm{q}}$ so that QA-ST can find a better assignment between sub-optimal ones.

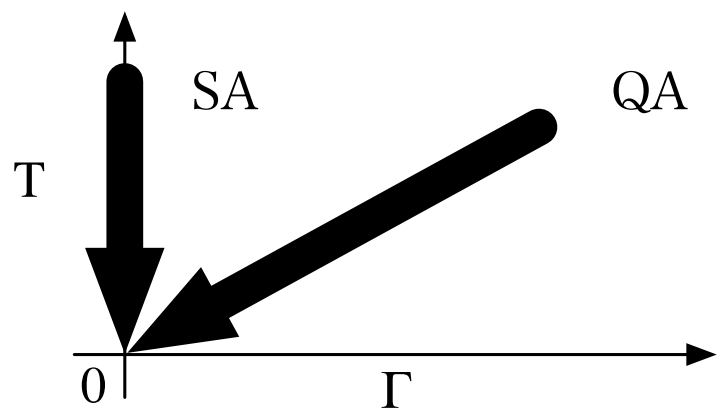

Figure 1: Quantum annealing (QA) adds another dimension to simulated annealing (SA) to control a model. QA iteratively decreases $T$ and $\Gamma$ whereas SA decreases just $T$.

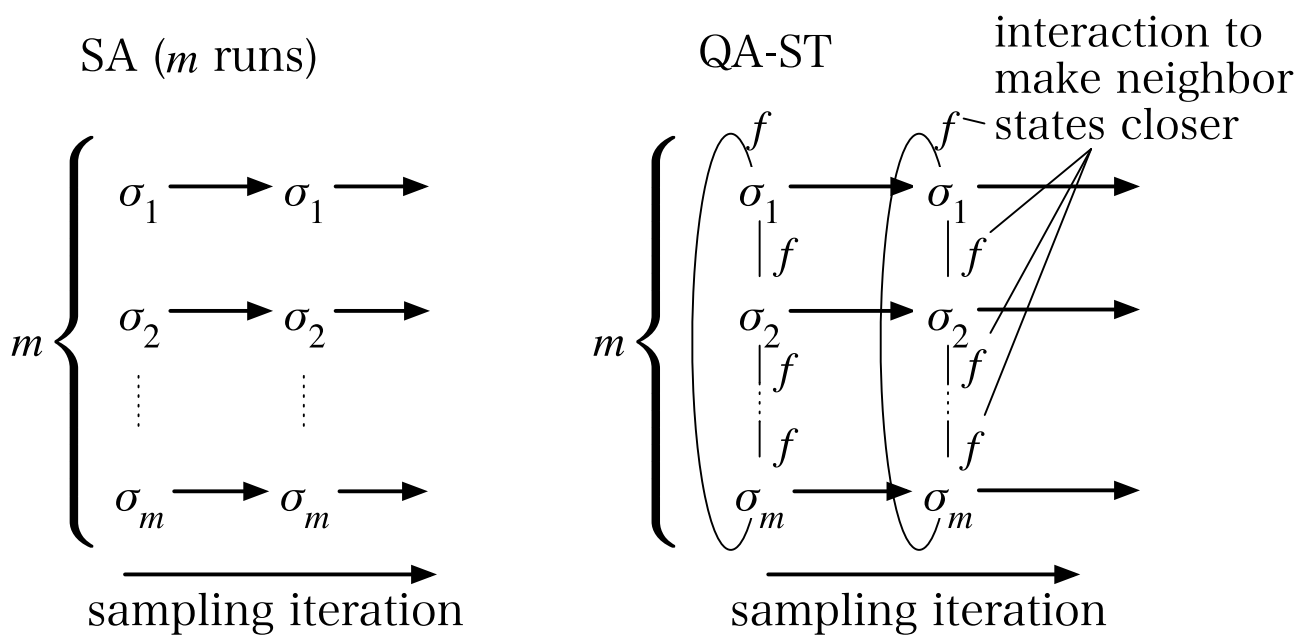

Figure 2: Illustrative explanation of QA. The left figure shows $m$ independent SAs, and the right one is QA algorithm derived with the Suzuki-Trotter (ST) expansion. $\sigma$ denotes a clustering assignment.

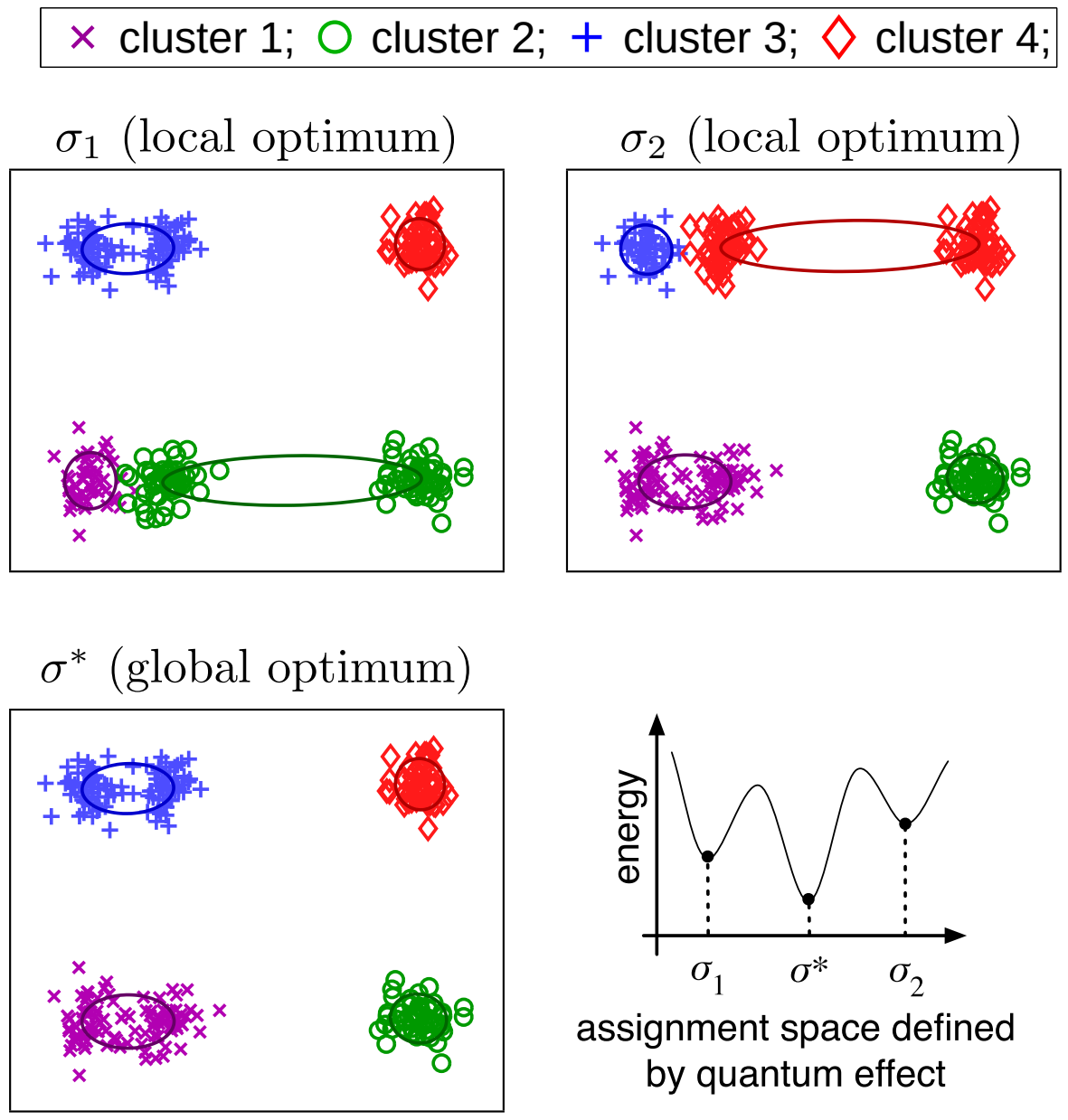

Figure 3: Three clustering results by a mixture of four Gaussians (i.e. #clusters=4).

## 2 Preliminaries

First of all, we introduce the notation used in this paper. We assume we have $n$ data points, and they are assigned to $k$ clusters. The assignment of the $i$-th data point is denoted by binary indicator vector $\tilde{\sigma}_i$. For example, when $k$ is equal to two, we denote the $i$-th data point assigned to the first and the second cluster by $\tilde{\sigma}_i = (1,\ 0)^T$ and $\tilde{\sigma}_i = (0,\ 1)^T$, respectively. The assignment of all data points is also denoted by an indicator vector, $\sigma$, whose length is $k^n$ because the number of available assignments is $k^n$. $\sigma$ is constructed with $\{\tilde{\sigma}_i\}_{i=1}^n$, $\sigma = \bigotimes_{i=1}^n \tilde{\sigma}_i$, where $\otimes$ is the Kronecker product, which is a special case of the tensor product for matrices. Let $A$ and $B$ be matrices where $A = \begin{pmatrix} a_{11} & a_{12} \\ a_{21} & a_{22} \end{pmatrix}$. Then, $A \otimes B = \begin{pmatrix} a_{11}B & a_{12}B \\ a_{21}B & a_{22}B \end{pmatrix}$ (see Minka (2000) for example). Only one element in $\sigma$ is one, and the others are zero. For example, $\sigma = \tilde{\sigma}_1 \otimes \tilde{\sigma}_2 = (0,\ 1,\ 0,\ 0)^T$ when $k=2, n=2$, the first data point is assigned to the first cluster ($\tilde{\sigma}_1 = (1,\ 0)^T$) and the second data point is assigned to the second cluster ($\tilde{\sigma}_2 = (0,\ 1)^T$). We also use $k$ by $n$ matrix $Y$ to denote the assignment of all data,

$$Y(\sigma) = (\tilde{\sigma}_1, \tilde{\sigma}_2, ..., \tilde{\sigma}_n). \tag{1}$$

We do not store $\sigma$ in memory whose length is $k^n$, but we store $Y$. We use $\sigma$ only for the derivation of quantum annealing. The proposed QA algorithm is like parallel $m$ SAs. We denote the $j$-th SA of the parallel SA by $\sigma_j$. The $i$-th data point in $\sigma_j$ is denoted by $\tilde{\sigma}_{j,i}$, s.t. $\sigma_j = \bigotimes_{i=1}^n \tilde{\sigma}_{j,i}$. When $A$ is a matrix, $e^A$ is the matrix exponential of $A$ defined by $e^A = \sum_{l=0}^{\infty} \frac{1}{l!} A^l$ .

## 3 Simulated Annealing for Clustering

We briefly review simulated annealing (SA) (Kirkpatrick et al., 1983) particularly for clustering. SA is a stochastic optimization algorithm. An objective function is given as an energy function such that a better solution has a lower energy. In each step, SA searches for the next random solution near the current one. The next solution is chosen with a probability that depends on temperature $T$ and on the energy function value of the next solution. SA almost ran-domly choose the next solution when $T$ is high, and it goes down the hill of the energy function when $T$ is low. Slower cooling $T$ increases the probability to find the global optimum.

Algorithm 1 summarizes a SA algorithm for clustering. Given inverse temperature $\beta = 1/T$, SA updates state $\sigma$ with,

$$p_{\mathrm{SA}}(\sigma;\beta) = \frac{1}{Z}\exp\left[-\beta E(\sigma)\right], \tag{2}$$

where $E(\sigma)$ is the energy function of state $\sigma$, and $Z$ is a normalization factor defined by $Z = \sum_\sigma \exp(-\beta E(\sigma))$. For probabilistic models, the energy function is defined by $E(\sigma) \equiv -\log p_{\text{prob-model}}(X,\sigma)$ where $p_{\text{prob-model}}(X,\sigma)$ is given by a probabilistic model and $X$ is data. Note $p_{\mathrm{SA}}(\sigma;\beta=1) = p_{\text{prob-model}}(\sigma|X)$. For loss-function-based models (e.g. spectral clustering), which searches for $\sigma = \operatorname{argmin}_\sigma loss(X,\sigma)$, the energy function is defined by $E(\sigma) = loss(X,\sigma)$.

In many cases, the calculation of $Z$ in (2) is intractable. Thus, Markov chain Monte Carlo (MCMC) is utilized to sample a new state from a current state. In this paper, we focus on the Gibbs sampler in MCMC methods. Each step of the Gibbs sampler draws an assignment of the $i$-th data point, $\tilde{\sigma}_i$, from,

$$p_{\mathrm{SA}}(\tilde{\sigma}_i|\sigma\backslash\tilde{\sigma}_i) = \frac{\exp\left[-\beta E(\sigma)\right]}{\sum_{\tilde{\sigma}_i}\exp\left[-\beta E(\sigma)\right]}, \tag{3}$$

where $\sigma\backslash\tilde{\sigma}_i$ means $\{\tilde{\sigma}_j|j \neq i\}$. Note the denominator of (3) is summation over $\tilde{\sigma}_i$, which is tractable ($\mathcal{O}(k)$).

## 4 Quantum Annealing for Clustering

Our goal of this section is to derive a sampling algorithm based on quantum annealing (QA) for clustering. On the way to the goal, our contribution is three folds, which are a well-formed quantum effect in Section 4.1, an appropriate similarity measure for clustering in Section 4.2 and an annealing schedule in Section 4.3.

### 4.1 Introducing Quantum Effect and the Suzuki-Trotter expansion

Before we apply QA to clustering problems, we set them up in a similar fashion to statistical mechanics. We reformulate (2) by the following equation,

$$p_{\mathrm{SA}}(\sigma;\beta) = \frac{1}{Z}\sigma^T e^{-\beta\mathcal{H}_{\mathrm{c}}}\sigma, \tag{4}$$

where $\mathcal{H}_{\mathrm{c}}$ is a $k^n$ by $k^n$ diagonal matrix. $\mathcal{H}_{\mathrm{c}}$ is called (classical) Hamiltonian in physics. For example, we have the following $\mathcal{H}_{\mathrm{c}}$ when $k=2$ and $n=2$.

$$\mathcal{H}_{\mathrm{c}} = \begin{pmatrix} E(\sigma^{(1)}) & 0 & 0 & 0 \\ 0 & E(\sigma^{(2)}) & 0 & 0 \\ 0 & 0 & E(\sigma^{(3)}) & 0 \\ 0 & 0 & 0 & E(\sigma^{(4)}) \end{pmatrix}. \tag{5}$$

In this example, $\sigma^{(t)}$ indicates the $t$-th assignment of $k^n$ available assignments, i,e. $\sigma^{(1)} = (1,\ 0,\ 0,\ 0)^T$, $\sigma^{(2)} = (0,\ 1,\ 0,\ 0)^T$, $\sigma^{(3)} = (0,\ 0,\ 1,\ 0)^T$ and $\sigma^{(4)} = (0,\ 0,\ 0,\ 1)^T$. $e^{-\beta\mathcal{H}_{\mathrm{c}}}$ is the matrix exponential in (4). Since $\mathcal{H}_{\mathrm{c}}$ is diagonal, $e^{-\beta\mathcal{H}_{\mathrm{c}}}$ is also diagonal with $[e^{-\beta\mathcal{H}_{\mathrm{c}}}]_{tt} = \exp(-\beta E(\sigma^{(t)}))$. Hence, we find $\sigma^{(t)T}e^{-\beta\mathcal{H}_{\mathrm{c}}}\sigma^{(t)} = \exp(-\beta E(\sigma^{(t)}))$ and (4) equal to (2). In practice, we use MCMC methods to sample $\sigma$ from $p_{\mathrm{SA}}(\sigma;\beta)$ in (4) by (3). This is because we do not need to calculate $Z$ and it is easy to evaluate $\sigma^T e^{-\beta\mathcal{H}_{\mathrm{c}}}\sigma$, which is equal to $\exp(-\beta E(\sigma))$.

QA draws a sample from the following equation,

$$p_{\mathrm{QA}}(\sigma;\beta,\Gamma) = \frac{1}{Z}\sigma^T e^{-\beta\mathcal{H}}\sigma, \tag{6}$$

where $\mathcal{H}$ is defined by $\mathcal{H} = \mathcal{H}_{\mathrm{c}} + \mathcal{H}_{\mathrm{q}}$. $\mathcal{H}_{\mathrm{q}}$ represents quantum effect. We define $\mathcal{H}_{\mathrm{q}}$ by $\mathcal{H}_{\mathrm{q}} = \sum_{i=1}^n \rho_i$ where

$$\rho_i = \left(\bigotimes_{j=1}^{i-1}\mathbb{E}_k\right)\otimes\rho\otimes\left(\bigotimes_{j=i+1}^{n}\mathbb{E}_k\right), \qquad \rho = \Gamma(\mathbb{E}_k - \mathbb{1}_k),$$

$\mathbb{E}_k$ is the $k$ by $k$ identity matrix, and $\mathbb{1}_k$ is the $k$ by $k$ matrix of ones, i.e. $[\mathbb{1}_k]_{ij} = 1$ for all $i$ and $j$. For example, $\mathcal{H}$ is,

$$\mathcal{H} = \begin{pmatrix} E(\sigma^{(1)}) & -\Gamma & -\Gamma & 0 \\ -\Gamma & E(\sigma^{(2)}) & 0 & -\Gamma \\ -\Gamma & 0 & E(\sigma^{(3)}) & -\Gamma \\ 0 & -\Gamma & -\Gamma & E(\sigma^{(4)}) \end{pmatrix}, \tag{7}$$

when $k=2$ and $n=2$. The derived algorithm depends on quantum effect $\mathcal{H}_{\mathrm{q}}$. We found our definition of $\mathcal{H}_{\mathrm{q}}$ worked well. We also tried a couple of $\mathcal{H}_{\mathrm{q}}$. We explain a bad example of $\mathcal{H}_{\mathrm{q}}$ later in this section.

QA samples $\sigma$ from (6). For SA, MCMC methods are exploited for sampling. However, in quantum models, we cannot apply MCMC methods directly to (6) because it is intractable to evaluate $\sigma^T e^{-\beta\mathcal{H}}\sigma$ unlike $\sigma^T e^{-\beta\mathcal{H}_{\mathrm{c}}}\sigma = \exp(-\beta E(\sigma))$. This is because $e^{-\beta\mathcal{H}}$ is not diagonal whereas $e^{-\beta\mathcal{H}_{\mathrm{c}}}$ is diagonal. Thus, we exploit the Trotter product formula (Trotter, 1959) to approximate (6). If $A_1, ..., A_L$ are symmetric matrices, the Trotter product formula gives, $\exp\left(\sum_{l=1}^L A_l\right) = \left(\prod_{l=1}^L \exp(A_l/m)\right)^m + O\left(\frac{1}{m}\right)$. Note the residual of finite $m$ is the order of $1/m$. Hence, this approximation becomes exact in the limit of $m \to \infty$. Since $\mathcal{H} = \mathcal{H}_c + \mathcal{H}_q$ is symmetric, we can apply the Trotter product formula to (6). Following (Suzuki, 1976), (6) reads the following expression,

**Theorem 4.1.**

$$p_{QA}(\sigma_1;\beta,\Gamma) = \sum_{\sigma_2}\cdots\sum_{\sigma_m} p_{QA\text{-}ST}(\sigma_1,\sigma_2,...,\sigma_m;\beta,\Gamma) + O\left(\frac{1}{m}\right), \quad (8)$$

*where*

$$p_{QA\text{-}ST}(\sigma_1,...,\sigma_m;\beta,\Gamma) \equiv \frac{1}{Z}\prod_{j=1}^{m} p_{SA}(\sigma_j;\beta/m)e^{s(\sigma_j,\sigma_{j+1})f(\beta,\Gamma)}, \quad (9)$$

$$s(\sigma_j,\sigma_{j+1}) = \frac{1}{n}\sum_{i=1}^{n}\delta(\tilde{\sigma}_{j,i},\tilde{\sigma}_{j+1,i}), \quad (10)$$

$$f(\beta,\Gamma) = n\log\left(1+\frac{k}{e^{\frac{k\beta\Gamma}{m}}-1}\right). \quad (11)$$

The derivation from (6) to (8) is called the Suzuki-Trotter expansion. We show the details of the derivation in Appendix A. (8) *means sampling* $\sigma_1$ *from* $p_{QA}(\sigma_1;\beta,\Gamma)$ *is approximated by sampling* $(\sigma_1,...,\sigma_m)$ *from* $p_{QA\text{-}ST}(\sigma_1,...,\sigma_m)$. (9) shows $p_{\text{QA-ST}}$ is similar to parallel $\{p_{\text{SA}}(\sigma_j;\beta/m)\}_{j=1}^m$, but it has quantum interaction $e^{s(\sigma_j,\sigma_{j+1})f(\beta,\Gamma)}$. Note if $f(\beta,\Gamma)=0$, i.e. $\Gamma=\infty$, the interaction disappears, and $p_{\text{QA-ST}}$ becomes $m$ independent SAs. $s(\sigma_j,\sigma_{j+1})$ takes $[0,1]$ where $s(\sigma_j,\sigma_{j+1})=1$ when $\sigma_j=\sigma_{j+1}$ and $s(\sigma_j,\sigma_{j+1})=0$ when $\sigma_j$ and $\sigma_{j+1}$ are completely different. Thus, we call $s(\sigma_j,\sigma_{j+1})$ similarity. Even with finite $m$, we can show the approximation in (8) becomes exact after enough annealing iterations has passed with our annealing schedule proposed in Section 4.3[1].

The similarity in (10) depends on quantum effect $\mathcal{H}_q$. A different $\mathcal{H}_q$ results in a different similarity. For example, we can derive an algorithm with quantum effect $\mathcal{H}'_q = \Gamma(\mathbb{E}_{k^n} - \mathbb{1}_{k^n})$. $\mathcal{H}'_q$ gives similarity $s'(\sigma_j,\sigma_{j+1}) = \prod_{i=1}^{n}\delta(\tilde{\sigma}_{j,i},\tilde{\sigma}_{j+1,i})$. Going back to Fig.3, we notice $s(\sigma_1,\sigma_2)>0$ but $s'(\sigma_1,\sigma_2)=0$. In this case, $p_{\text{QA-ST}}$ with $\mathcal{H}'_q$ is just $m$ independent SAs because interaction $f$ is canceled by $s'(\sigma_1,\sigma_2)=0$, and $p_{\text{QA-ST}}$ is unlikely to search for $\sigma^*$. On the other hand, $p_{\text{QA-ST}}$ with $\mathcal{H}_q$ is more likely to search for $\sigma^*$ because interaction $f$ allows $\sigma_1$ and $\sigma_2$ to go between $\sigma_1$ and $\sigma_2$.

Now, we can construct a Gibbs sampler based on $p_{\text{QA-ST}}$ in a similar fashion to (3). Although the sampler is tractable for statistical mechanics, it is intractable for machine learning. We give a solution to the problem in Section 4.2. We also discuss the annealing schedule of $\beta$ and $\Gamma$ in Section 4.3, which is a crucial point in practice.

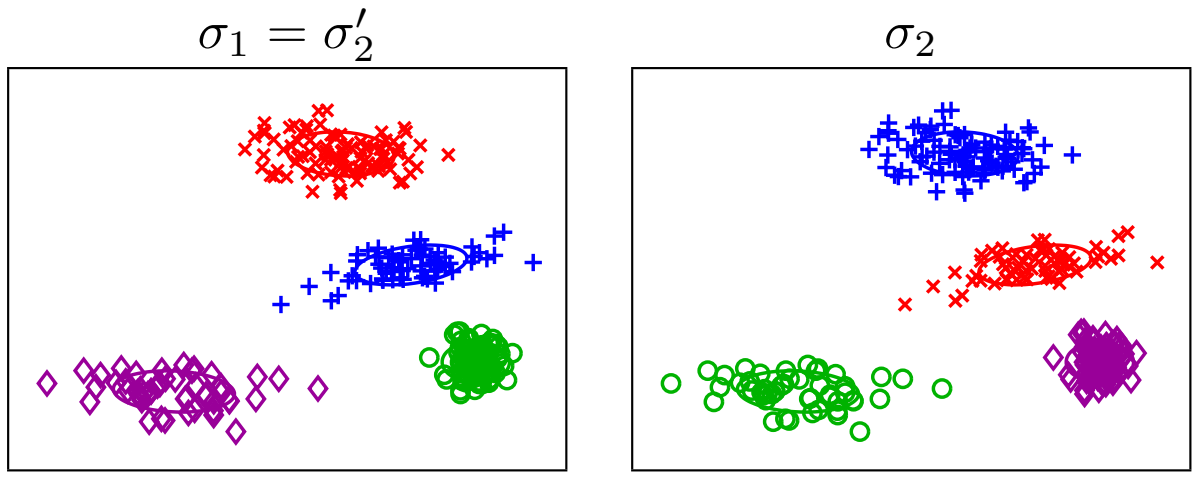

Figure 4: $\sigma_1$ and $\sigma_2$ give the same clustering but have different cluster labels. Thus, $s(\sigma_1,\sigma_2)=0$. After cluster label permutation from $\sigma_2$ to $\sigma'_2$, $s(\sigma_1,\sigma'_2)=1$. The *purity*, $\tilde{s}$, gives $\tilde{s}(\sigma_1,\sigma_2)=1$ as well.

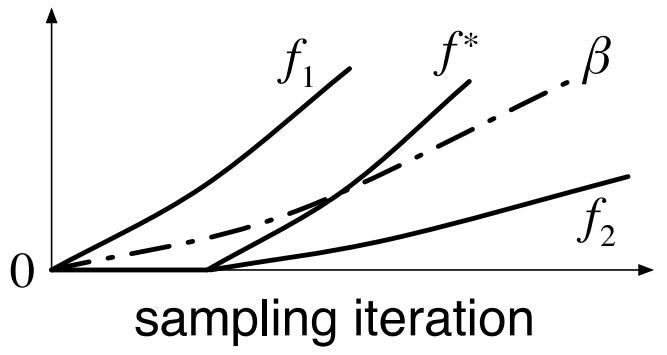

Figure 5: The schedules of $\beta$ and $f(\beta,\Gamma)$.

## 4.2 Cluster-Label Permutation

Our goal is to make an efficient sampling algorithm. In a similar fashion to (3), we can construct a Gibbs sampler $p_{\text{QA-ST}}(\tilde{\sigma}_{j,i}|\{\sigma\}_{j=1}^m\backslash\tilde{\sigma}_{j,i})$ whose computational complexity is the same as that of (3). However, the sampler can easily get stuck in local optima, which is for example $p_{\text{QA-ST}}(\sigma_1,\sigma_2)$ in Fig.4. If we can draw $\sigma'_2$ in Fig.4 from $\sigma_2$, $p_{\text{QA-ST}}(\sigma_1,\sigma'_2)$ is a better state than $p_{\text{QA-ST}}(\sigma_1,\sigma_2)$ i.e. $p_{\text{QA-ST}}(\sigma_1,\sigma'_2) \geq p_{\text{QA-ST}}(\sigma_1,\sigma_2)$ because $s(\sigma_1,\sigma'_2)=1$, $s(\sigma_1,\sigma_2)=0$ and $f(\beta,\Gamma)\geq 0$ in (9). Since sampler $p_{\text{QA-ST}}(\tilde{\sigma}_{j,i}|\{\sigma\}_{j=1}^m\backslash\tilde{\sigma}_{j,i})$ only changes the label of one data point at a time, the sampler cannot sample $\sigma'_2$ from $\sigma_2$ efficiently. In statistical mechanics, a cluster label permutation sampler is applied to cases such as Fig.4. The label permutation sampler does not change cluster assignments but draws cluster label permutation, e.g. $\sigma'_2$ from $\sigma_2$ in one step. In other words, the sampler exchanges rows of matrix $Y(\sigma)$ defined in (1). In the case of Fig.4, $k$ is equal to four, so we have $4!=24$ choices of label permutation. The computational complexity of the sampler is $\mathcal{O}(k!)$ because its normalization factor requires summation over $k!$ choices. The sampler is tractable for statistical mechanics due to relatively small $k$. However, it is intractable for machine learning where $k$ can be very large.

We introduce approximation of $p_{\text{QA-ST}}$ so that we do not need to sample label permutation, whose com-

[1]The residual of the approximation in (8) is dominated by $\beta^2\Gamma$ with small $\beta$ and large $\Gamma$. Using the annealing schedule proposed in Section 4.3, the residual goes to zero as annealing continues ($\beta\to 0$ and $\Gamma\to\infty$).

**Algorithm 1** Simulated Annealing for Clustering

1: Initialize inverse temperature $\beta$ and assignment $\sigma$.
2: **repeat**
3:   **for** $i = 1, ..., n$ **do**
4:     Draw the new assignment of the $i$-th data point, $\tilde{\sigma}_i$, with a probability given in (3).
5:   **end for**
6:   Increase inverse temperature $\beta$.
7: **until** State $\sigma$ converges

putational complexity is $\mathcal{O}(k!)$. In particular, we replace similarity $s(\sigma_j, \sigma_{j+1})$ in (9) by the *purity*, $\tilde{s}(\sigma_j, \sigma_{j+1})$. The *purity*, $\tilde{s}(\sigma_j, \sigma_{j+1})$, is defined by $\tilde{s}(\sigma_j, \sigma_{j+1}) \equiv \frac{1}{n} \sum_{c=1}^{k} \max_{c'=1...k} \left[ Y(\sigma_j) Y(\sigma_{j+1})^T \right]_{c,c'}$ where $Y$ is defined in (1), and $[A]_{c,c'}$ denotes the $(c, c')$ element of matrix $A$. In the case of Fig.4, $\tilde{s}(\sigma_1, \sigma_2) = 1$ whereas $s(\sigma_1, \sigma_2) = 0$. In general, $s(\sigma_1, \sigma_2) \leq \tilde{s}(\sigma_1, \sigma_2)$.

Let $\tilde{\sigma}_{j,i}$ be the $i$-th data point of assignment $\sigma_j$. The update probability of $\tilde{\sigma}_{j,i}$ with the *purity* is,

$$p_{\text{QA-ST}+purity}(\tilde{\sigma}_{j,i} | \{\sigma_j\}_{j=1}^m \backslash \tilde{\sigma}_{j,i}; \beta, \Gamma) = \frac{\exp\left[ -\frac{\beta}{m} E(\sigma_j) + \tilde{s}(\sigma_{j-1}, \sigma_j, \sigma_{j+1}) f(\beta, \Gamma) \right]}{\sum_{\tilde{\sigma}_{j,i}} \exp\left[ -\frac{\beta}{m} E(\sigma_j) + \tilde{s}(\sigma_{j-1}, \sigma_j, \sigma_{j+1}) f(\beta, \Gamma) \right]}, \quad (12)$$

where $\tilde{s}(\sigma_{j-1}, \sigma_j, \sigma_{j+1}) = \tilde{s}(\sigma_j, \sigma_{j-1}) + \tilde{s}(\sigma_j, \sigma_{j+1})^2$. The computational complexity of (12) is $\mathcal{O}(k^2)$, but caching statistics reduces it to $\mathcal{O}(k)$. Thus, Step 1 in Algorithm 2 requires $\mathcal{O}(k^2)$, and Step 5 requires $\mathcal{O}(k)$, which is the same as SA.

Using another representation of $\sigma_j$ and a different $\mathcal{H}_\text{q}$, we can develop a sampler, which does not need label permutation. However, its computational complexity is $\mathcal{O}(n)$ in Step 5, which is much more expensive than $\mathcal{O}(k)$. Thus, the sampler is less efficient than the proposed sampler even though the sampler does not need to solve label permutation.

### 4.3 Annealing Schedule of $\beta$ and $\Gamma$

The annealing schedules of $\beta$ and $\Gamma$ significantly affect the result of QA. Thus, it is crucial to use good schedules of $\beta$ and $\Gamma$. In this section, we propose the annealing schedule of $\Gamma$ and $\beta$.

We address two points before proposing a schedule. One is our observation from pilot experiments, and the other is the balance of $\beta$ and $\Gamma$. From our pilot

**Algorithm 2** Quantum Annealing for Clustering

1: Initialize inverse temperature $\beta$ and quantum annealing parameter $\Gamma$.
2: **repeat**
3:   **for** $j = 1, ..., m$ **do**
4:     **for** $i = 1, ..., n$ **do**
5:       Draw the new assignment of the $i$-th data point, $\sigma_{j,i}$, with a probability given in (12).
6:     **end for**
7:   **end for**
8:   Increase inverse temperature $\beta$, and decrease QA parameter $\Gamma$.
9: **until** State $\sigma$ converges

experiments, we observe QA-ST works well when it can find suboptimal assignments $\{\sigma_j\}_{j=1}^m$ by convergence. (12) shows QA-ST searches for a better assignment from suboptimal $\{\sigma_j\}_{j=1}^m$. On the other hand, when current $\{\sigma_j\}_{j=1}^m$ are far away from global optimum or even sub-optima, QA-ST does not necessarily work well. Comparing (12) with (3) in terms of $\beta$ and $\Gamma$, if $\frac{\beta}{m} \gg f(\beta, \Gamma)$, $\{\sigma_j\}_{j=1}^m$ are sampled from $p_{\text{SA}}(\sigma_j)$, i.e. no interaction between $\sigma_j$ and $\sigma_{j+1}$. On the other hand, if $\frac{\beta}{m} \ll f(\beta, \Gamma)$, $\{\sigma_j\}_{j=1}^m$ become very close to each other regardless of energy $E(\sigma_j)$.

From the above discussion, $\beta/m$ at the beginning should be larger than $f(\beta, \Gamma)$ and large enough to collect suboptimal assignments, and $f(\beta, \Gamma)$ should become larger than $\beta/m$ at some point to make $\{\sigma_j\}_{j=1}^m$ closer. The best path of $\beta$ and $f(\beta, \Gamma)$ would be like $f^*$ in Fig.5. $f_1$ in Fig.5 is stronger than $\beta$ from the beginning, which does not allow QA-ST to search for good assignments due to too strong quantum interaction $f(\beta, \Gamma)\tilde{s}(\sigma_{j-1}, \sigma_j, \sigma_{j+1})$ in (12). $f_2$ is always smaller than $\beta$. Hence, QA-ST never makes $\{\sigma_j\}_{j=1}^m$ closer. In other words, QA-ST does not search for a middle (hopefully better) assignment from $\{\sigma_j\}_{j=1}^m$. Specifically, we use the following annealing schedule in Step 8 in Algorithm 2.

$$\beta = \beta_0 r_\beta^i, \qquad \Gamma = \Gamma_0 \exp(-r_\Gamma^i), \quad (13)$$

where $r_\beta$ and $r_\Gamma$ are constants and $i$ denotes the $i$-th iteration of sampling. (13) comes from the following analysis of $f(\beta, \Gamma)$. When $\frac{k\beta\Gamma}{m} \ll 1$, (11) reads, $f(\beta, \Gamma) \approx -n \log\left(\frac{\beta\Gamma}{m}\right) = n r_\Gamma^i - n \log\left(\frac{\beta\Gamma_0}{m}\right)$ . Thus, the path of $f(\beta, \Gamma)$ become $f^*$ in Fig.5 when $\Gamma_0$ is large enough and $r_\beta < r_\Gamma$. In this paper[3], we set $\Gamma_0$ to a large value such that $f(\beta, \Gamma) \approx 0$ until $\beta = m$. This means $p_{\text{QA-ST}}(\sigma_1, ..., \sigma_m)$ is just $m$ independent

[2]Note $\tilde{s}$ is not commutative, and take care of the order of the arguments of $\tilde{s}$. We use $\tilde{s}(\sigma_{j-1}, \sigma_j, \sigma_{j+1}) = \tilde{s}(\sigma_j, \sigma_{j-1}) + \tilde{s}(\sigma_j, \sigma_{j+1})$, but we omit the reason due to space.

[3]When QA-ST is applied to loss-function-based models, "until $\beta = m$" should be calibrated according to loss-functions.

instances of $p_{\mathrm{SA}}(\sigma_j;\beta/m)$ until $\beta = m$.

Note there is not much difference of difficulty between SA and QA-ST to choose annealing schedules. In general, we should choose the schedule of $\Gamma$ to be $f^*$ when the schedule of $\beta$ is given. As shown in the next section, QA-ST works well with $r_\Gamma \approx r_\beta \times 1.05$. Thus, the difficulty of choosing annealing schedules for QA-ST is reduced to that of choosing the schedule of $\beta$ for SA.

## 5 Experiments

We show experimental results in Fig.6. In the top three rows of Fig.6, we vary the schedule of $\Gamma$ with a fixed schedule of $\beta$ to see QA-ST work better than the best energy of $m$ SAs when the schedule of $\Gamma$ lets the path of $f(\beta,\Gamma)$ be $f^*$ in Fig.5. In the bottom row of the figure, we compare QA-ST and SA with a slower schedule of $\beta$. This experiment shows whether QA-ST still works better than SA or not while the slower schedule of $\beta$ improves SA.

We apply SA and QA-ST to two models, which are a mixture of Gaussians (MoG) with a conjugate normal-inverse-Wishart prior and the latent Dirichlet allocation (LDA) (Blei et al., 2003). For both models, parameters are marginalized out, and $E(\sigma) \equiv -\log p(X,\sigma)$ where $X$ is data. Thus, QA-ST and SA search for maximum a posteriori (MAP) assignment $\sigma$. MoG is applied to MNIST data set, and LDA is applied to NIPS corpus and Reuters. For MNIST, we randomly choose 5,000 data points and apply PCA to reduce the dimensionality to 20. NIPS corpus has 1,684 documents, and we randomly choose 1000 words in vocabulary. We also randomly choose 1000 documents and 2000 words in vocabulary from Reuters. We set $k$ to 30, 20 and 20 for MNIST, NIPS corpus and Reuters, respectively. We use the same schedule of $\beta$ for QA-ST and SA. In particular, we use the same $r_\beta$ for QA-ST and SA, and we set $\beta_0 = 0.2$ for SA and $\beta_0 = 0.2m$ for QA-ST. The difference of $\beta_0$ for SA and QA-ST keeps QA-ST similar to SA in terms of $\beta$-annealing for fair comparison (see (3) and (12)). For each data, we vary $r_\Gamma$ from 1.02 to 1.20 with fixed $r_\beta$. When $r_\beta \le r_\Gamma$, the path of $f(\beta,\Gamma)$ becomes $f^*$ in Fig.5. For any data set, $m$ is set to 50 for QA-ST. We set $m$ of SAs so that $m$ SAs consume the same time as QA-ST. Thus, we can compare QA-ST and $m$ SAs in terms of iteration in Fig.6. Consequently, $m$ of SA was set to 51, 55 and 55 for MNIST, NIPS and Reuters, respectively[4]. In Fig.6, we plot only after $\beta = m$ for QA-ST and $\beta = 1$ for SA, which happen at the same iteration for QA-ST and SA.

[4]QA-ST and $m$ SAs took 21.7 and 22.0 hours for MNIST, 62.5 and 62.8 hours for NIPS and 9.9 and 10.0 hours for Reuters.

In Fig.6, the left column and the middle column show the minimum and the mean energy of $\{\sigma_j\}_{j=1}^m$. Since this is an optimization problem, we are interested in the minimum energy in the left column. For each data, QA-ST with $f^*$ achieved better results than SA. The right column of Fig.6 shows the mean of purity $\tilde{s}$. As we expect, the larger $r_\Gamma$ resulted in the higher $\tilde{s}$. The bottom row of Fig.6 shows the result of NIPS with the slower schedule of $\beta$ than the schedule in the third row of Fig.6. Although SA found better results than the third row of Fig.6, QA-ST still worked better than SA. Our experimental results are also consistent with the claim of Matsuda et al. (2009), which is that QA works the better than SA for more difficult problems. QA worked better for LDA than MoG. The right column of Fig.6 shows $\tilde{s}$ of LDA converged to smaller values than that of MoG. This means LDA has more local optima than MoG.

In this section, we have shown QA-ST achieves better results than SA when the schedule of $\Gamma$ is $f^*$ in Fig.5. We have also shown even with the slower schedule of $\beta$, QA-ST still works better than SA.

## 6 Discussion & Conclusion

Many techniques to accelerate sampling have been studied. Such techniques can be applied to the proposed algorithm. For example, the split-merge sampler (Richardson and Green, 1997) and the permutation augmented sampler (Liang et al., 2007) use a global move to escape from local minima. These techniques are available for the proposed algorithm as well. We can also apply the exchange Monte Carlo method.

We have applied quantum annealing (QA) to clustering. To our best knowledge, this is the first study of QA for clustering. We have proposed quantum effect $\mathcal{H}_{\mathrm{q}}$ fit to clustering and derived a QA-based sampling algorithm. We have also proposed a good annealing schedule for QA, which is crucial for applications. The computational complexity of QA is larger than a single simulated annealing (SA). However, we have empirically shown QA finds a better clustering assignment than the best one of multiple-run SAs that are randomly restarted until they consumes the same time as QA. In other words, QA is better than SA when we run SA many times. Actually, it is typical to run SA many times because SA's fast cooling schedule of temperature $T$ does not necessarily find the global optimum. Thus, we strongly believe QA is a novel alternative to SA for optimizing clustering. In addition, it is easy to implement the proposed algorithm because it is very similar to SA.

Unfortunately, there is no proof yet that QA is better than SA in general. Thus, we need to experimen-tally show QA's performance for each problem like this paper. However, it is worth trying to develop QA-based algorithms for different models, e.g. Bayesian networks, by different quantum effect $\mathcal{H}_q$. The proposed algorithm looks like genetic algorithms in terms of running multiple instances. Studying their relationship is also interesting future work.

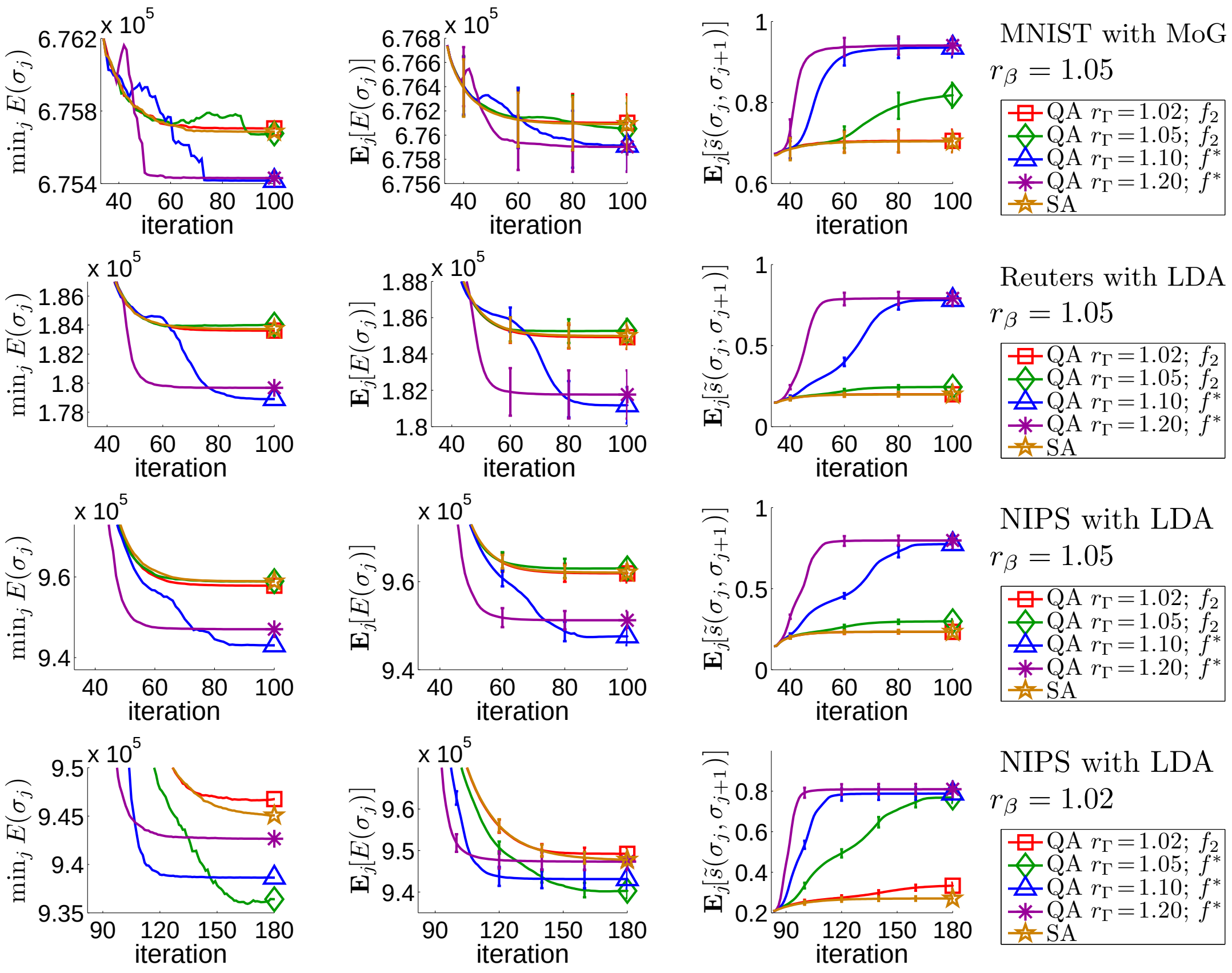

Figure 6: Comparison between SA and QA varying annealing schedule. $r_\Gamma$, $f_2$ and $f^*$ in legends correspond to Fig.5. The left-most column shows what SA and QA found. QA with $f^*$ always found better results than SA.

### Acknowledgements

This work was partially supported by Research on Priority Areas "Physics of new quantum phases in superclean materials" (Grant No. 17071011) from MEXT, and also by the Next Generation Super Computer Project, Nanoscience Program from MEXT. Special thanks to Taiji Suzuki, T-PRIMAL members and Sato Lab.

# A The Details of the Suzuki-Trotter Expansion

Following Suzuki (1976), we give the details of derivation of Theorem 4.1. The following Trotter product formula (Trotter, 1959) says if $A_1, \cdots, A_n$ are symmetric matrices, we have

$$\exp\left(\sum_{l=1}^{L} A_l\right) = \left(\prod_{l=1}^{L} \exp(A_l/m)\right)^m + O\left(\frac{1}{m}\right). \quad (14)$$

Applying the Trotter product formula to (6), we have

$$\begin{aligned} p_{\mathrm{QA}}(\sigma_1;\beta,\Gamma) &= \frac{1}{Z}\sigma_1^T e^{-\beta(\mathcal{H}_\mathrm{c}+\mathcal{H}_\mathrm{q})}\sigma_1 \\ &= \frac{1}{Z}\sigma_1^T \left(e^{-\frac{\beta}{m}\mathcal{H}_\mathrm{c}} e^{-\frac{\beta}{m}\mathcal{H}_\mathrm{q}}\right)^m \sigma_1 + O\left(\frac{1}{m}\right). \quad (15) \end{aligned}$$

Note

$$\begin{aligned} \sigma_1^T \left(e^A\right)^2 \sigma_1 = \sigma_1^T e^A \mathbb{E}_k e^A \sigma_1 &= \sigma_1^T e^A \left(\sum_{\sigma_2} \sigma_2 \sigma_2^T\right) e^A \sigma_1 \\ &= \sum_{\sigma_2} \sigma_1^T e^A \sigma_2 \sigma_2^T e^A \sigma_1. \quad (16) \end{aligned}$$

Hence, we express (15) by marginalizing out auxiliary variables $\{\sigma_1', \sigma_2, \sigma_2', ..., \sigma_m, \sigma_m'\}$,

$$\begin{aligned} &\frac{1}{Z}\sigma_1^T \left(e^{-\frac{\beta}{m}\mathcal{H}_\mathrm{c}} e^{-\frac{\beta}{m}\mathcal{H}_\mathrm{q}}\right)^m \sigma_1 \quad (17) \\ &= \frac{1}{Z}\sum_{\sigma_1'}\sum_{\sigma_2} \cdots \sum_{\sigma_m}\sum_{\sigma_m'} \sigma_1^T e^{-\frac{\beta}{m}\mathcal{H}_\mathrm{c}} \sigma_1' \sigma_1'^T e^{-\frac{\beta}{m}\mathcal{H}_\mathrm{q}} \sigma_2 \times \ldots \\ &\qquad \times \sigma_m^T e^{-\frac{\beta}{m}\mathcal{H}_\mathrm{c}} \sigma_m' \sigma_m'^T e^{-\frac{\beta}{m}\mathcal{H}_\mathrm{q}} \sigma_{m+1} \quad (18) \\ &= \frac{1}{Z}\sum_{\sigma_1'}\sum_{\sigma_2} \cdots \sum_{\sigma_m}\sum_{\sigma_m'} \prod_{j=1}^{m} \sigma_j^T e^{-\frac{\beta}{m}\mathcal{H}_\mathrm{c}} \sigma_j' \sigma_j'^T e^{-\frac{\beta}{m}\mathcal{H}_\mathrm{q}} \sigma_{j+1}, \quad (19) \end{aligned}$$

where $\sigma_{m+1} = \sigma_1$. To simplify (19) more, we use the following Lemma A.1 and Lemma A.2.

**Lemma A.1.**

$$\begin{aligned} \sigma_j^T e^{-\frac{\beta}{m}\mathcal{H}_\mathrm{c}} \sigma_j' &= \exp\left(-\frac{\beta}{m}E(\sigma_j)\right)\delta(\sigma_j, \sigma_j') \\ &\propto p_{SA}(\sigma_j;\beta/m)\delta(\sigma_j,\sigma_j'), \quad (20) \end{aligned}$$

*where $\delta(\sigma_j, \sigma_j') = 1$ if $\sigma_j = \sigma_j'$ and $\delta(\sigma_j, \sigma_j') = 0$ otherwise.*

*Proof.* By the definition, $e^{-\frac{\beta}{m}\mathcal{H}_\mathrm{c}}$ is diagonal with $[e^{-\frac{\beta}{m}\mathcal{H}_\mathrm{c}}]_{tt} = E(\sigma^{(t)})$, and $\sigma_j$ and $\sigma_j'$ are binary indicator vectors, i.e. only one element in $\sigma_j$ is one and the others are zero. Thus, the above lemma holds. □

**Lemma A.2.**

$$\sigma_j'^T e^{-\frac{\beta}{m}\mathcal{H}_\mathrm{q}} \sigma_{j+1} \propto e^{s(\sigma_j', \sigma_{j+1}) f(\beta,\Gamma)}. \quad (21)$$

*Proof.* Substituting $(A \otimes B)(C \otimes D) = (AC) \otimes (BD)$ and $e^{A_1 + A_2} = e^{A_1} e^{A_2}$ when $A_1 A_2 = A_2 A_1$, we find,

$$\begin{aligned} \sigma_j'^T e^{-\frac{\beta}{m}\mathcal{H}_\mathrm{q}} \sigma_{j+1} &= \sigma_j'^T \left(\bigotimes_{i=1}^{n} e^{-\frac{\beta}{m}\rho_i}\right)\sigma_{j+1} \\ &= \prod_{i=1}^{n} \tilde{\sigma}_{j,i}'^T e^{-\frac{\beta}{m}\rho} \tilde{\sigma}_{j+1,i}, \quad (22) \end{aligned}$$

where $\tilde{\sigma}_{j,i}$ is the $i$-th element of Kronecker product of $\sigma_j$, s.t. $\sigma_j = \bigotimes_{i=1}^{n} \tilde{\sigma}_{j,i}$. Substituting the following (23) and (24) into (22),

$$e^{-\frac{\beta}{m}\rho} = \sum_{l=0}^{\infty} \frac{1}{l!}\left(-\frac{\beta}{m}\right)^l \rho^l \quad (23)$$

$$\begin{aligned} \rho^l &= \Gamma^l(\mathbb{E}_k - \mathbb{1}_k)^l = \Gamma^l\left\{\sum_{i=0}^{l} \binom{l}{i} \mathbb{E}_k^i (-\mathbb{1}_k)^{l-i}\right\} \\ &= \Gamma^l\left\{\mathbb{E}_k + \sum_{i=0}^{l-1} \binom{l}{i} k^{l-i-1}(-1)^{l-i}\mathbb{1}_k\right\} \\ &= \Gamma^l\left\{\mathbb{E}_k + \frac{1}{k}\sum_{i=0}^{l-1} \binom{l}{i} (-k)^{l-i}\mathbb{1}_k\right\} \\ &= \Gamma^l\left\{\mathbb{E}_k + \frac{1}{k}\left\{(1-k)^l - 1\right\}\mathbb{1}_k\right\}, \quad (24) \end{aligned}$$

we have

$$\begin{aligned} &\sigma_j'^T e^{-\frac{\beta}{m}\mathcal{H}_\mathrm{q}} \sigma_{j+1} \\ &= \prod_{i=1}^{n}\sum_{l=0}^{\infty} \frac{1}{l!}\left(-\frac{\beta\Gamma}{m}\right)^l \tilde{\sigma}_{j,i}'^T \left(\mathbb{E}_k + \frac{1}{k}\left\{(1-k)^l - 1\right\}\mathbb{1}_k\right)\tilde{\sigma}_{j+1,i} \\ &= \prod_{i=1}^{n}\sum_{l=0}^{\infty} \frac{1}{l!}\left(-\frac{\beta\Gamma}{m}\right)^l \left\{\delta(\tilde{\sigma}_{j,i}', \tilde{\sigma}_{j+1,i}) + \frac{1}{k}(1-k)^l - \frac{1}{k}\right\} \\ &= \prod_{i=1}^{n}\left\{e^{-\frac{\beta\Gamma}{m}}\delta(\tilde{\sigma}_{j,i}', \tilde{\sigma}_{j+1,i}) + \frac{1}{k}e^{-\frac{\beta\Gamma}{m}(1-k)} - \frac{1}{k}e^{-\frac{\beta\Gamma}{m}}\right\} \\ &\propto e^{s(\sigma_j', \sigma_{j+1}) f(\beta,\Gamma)}. \quad (25) \end{aligned}$$

□

Using Lemma A.1 and Lemma A.2 into (19), (17) becomes,

$$\begin{aligned} &\frac{1}{Z}\sigma_1^T \left(e^{-\frac{\beta}{m}\mathcal{H}_\mathrm{c}} e^{-\frac{\beta}{m}\mathcal{H}_\mathrm{q}}\right)^m \sigma_1 \\ &= \frac{1}{Z}\sum_{\sigma_2} \cdots \sum_{\sigma_m} \prod_{j=1}^{m} p_{\mathrm{SA}}(\sigma_j;\beta/m) e^{s(\sigma_j,\sigma_{j+1}) f(\beta,\Gamma)}. \end{aligned}$$

From (15) and the above expression, we have shown Theorem 4.1.